\documentclass[conference]{IEEEtran}
\IEEEoverridecommandlockouts

\usepackage{cite}
\usepackage{amsmath,amssymb,amsfonts}
\usepackage{algorithmic}
\usepackage{graphicx}
\usepackage{textcomp}

\usepackage{booktabs}
\usepackage[table,xcdraw]{xcolor} 
\usepackage{caption}
\usepackage{array} 
\usepackage{multirow}
\usepackage{array}

\usepackage{makecell}
\usepackage{hyperref} 
\hypersetup{colorlinks=false}
\AtBeginDocument{\hypersetup{pdfborder={0 0 0}}}
\usepackage{xurl} 

\def\BibTeX{{\rm B\kern-.05em{\sc i\kern-.025em b}\kern-.08em
    T\kern-.1667em\lower.7ex\hbox{E}\kern-.125emX}}
\begin{document}

\title{Automated Detection of Malignant Lesions in the Ovary Using Deep Learning Models and XAI}

\author{
\IEEEauthorblockN{\hspace{-2mm}Md. Hasin Sarwar Ifty}
\IEEEauthorblockA{\hspace{-7mm}\textit{Dept. of Computer Science and Engr.} \\
\textit{\hspace{-7mm}BRAC University}\\
\hspace{-7mm}Dhaka, Bangladesh \\
\hspace{-2mm}hasin.sarwar.ifty@g.bracu.ac.bd}
\and
\IEEEauthorblockN{\hspace{-2mm}Nisharga Nirjan}
\IEEEauthorblockA{\textit{Dept. of Computer Science and Engr.} \\
\textit{\hspace{-7mm}BRAC University}\\
\hspace{-7mm}Dhaka, Bangladesh \\
\hspace{-2mm}nisharga.nirjan@g.bracu.ac.bd}
\and
\IEEEauthorblockN{Labib Islam}
\IEEEauthorblockA{\textit{Dept. of Computer Science and Engr.} \\
\textit{BRAC University}\\
Dhaka, Bangladesh \\
labib.islam@g.bracu.ac.bd}
\and
\IEEEauthorblockN{M. A. Diganta}
\IEEEauthorblockA{\textit{Dept. of Computer Science and Engr.} \\
\textit{BRAC University}\\
Dhaka, Bangladesh \\
m.a.diganta@g.bracu.ac.bd}
\and
\IEEEauthorblockN{Reeyad Ahmed Ornate}
\IEEEauthorblockA{\textit{Dept. of Computer Science and Engr.} \\
\textit{BRAC University}\\
Dhaka, Bangladesh \\
reeyad.ahmed.ornate@g.bracu.ac.bd}
\and
\IEEEauthorblockN{Anika Tasnim}
\IEEEauthorblockA{\textit{Dept. of Computer Science and Engr.} \\
\textit{BRAC University}\\
Dhaka, Bangladesh \\
ext.anika.tasnim@bracu.ac.bd}
\and
\IEEEauthorblockN{Md. Saiful Islam}
\IEEEauthorblockA{\textit{Dept. of Computer Science and Engr.} \\
\textit{BRAC University}\\
Dhaka, Bangladesh \\
md.saiful.islam@bracu.ac.bd}
}

\maketitle

\begin{abstract}
  The unrestrained proliferation of cells that are malignant in nature is cancer. In recent times, medical professionals are constantly acquiring enhanced diagnostic and treatment abilities by implementing deep learning models to analyze medical data for better clinical decision, disease diagnosis and drug discovery. A majority of cancers are studied and treated by incorporating these technologies. However, ovarian cancer remains a dilemma as it has inaccurate non-invasive detection procedures and a time consuming, invasive procedure for accurate detection. Thus, in this research, several Convolutional Neural Networks such as LeNet-5, ResNet, VGGNet and GoogLeNet/Inception have been utilized to develop 15 variants and choose a model that accurately detects and identifies ovarian cancer. For effective model training, the dataset OvarianCancer\&SubtypesDatasetHistopathology from Mendeley has been used. After constructing a model, we utilized Explainable Artificial Intelligence (XAI) models such as LIME, Integrated Gradients and SHAP to explain the black box outcome of the selected model. For evaluating the performance of the model, Accuracy, Precision, Recall, F1-Score, ROC Curve and AUC have been used. From the evaluation, it was seen that the slightly compact InceptionV3 model with ReLu had the overall best result achieving an average score of 94\% across all the performance metrics in the augmented dataset. Lastly for XAI, the three aforementioned XAI have been used for an overall comparative analysis. It is the aim of this research that the contributions of the study will help in achieving a better detection method for ovarian cancer.
\end{abstract}


\section{Introduction}
Cancer refers to a condition where some cells within the body grow uncontrollably, that is, the cells proliferate without any form of instruction from the body \cite{b30}. Malignant tumors or neoplasms are oftentimes correlated with cancer. One of the defining features of cancer is its ability to expand outside of its normal boundaries \cite{b10}. That is, a cancer affected part can create abnormal cells that can spread to other parts of the body and create cancerous cells there. This process is known as metastasis \cite{b10}. The primary cause of death due to cancer is metastasis. According to World Cancer Research Fund International \cite{b11} or WCRF International in short, the number of new cancer cases in the year 2020 was 18.1 million globally. A majority of cancers can be detected in early or middle stages and be treated effectively. However, there are cancers that cannot be detected until their advanced stage and thus, makes treatment of said cancers much harder. Ovarian cancer is one such cancer that is detected at its advanced stage only \cite{b12}. This cancer refers to abnormal growth of tumors in the ovaries. This makes it most lethal to women as it has no screening tests \cite{b13}. Many of the other cancers common among women such as Breast Cancer, Cervical Cancer can be detected via specialized tests. Mammograms and CBEs (Clinical Breast Exam) are commonly performed to detect Breast Cancer, whereas a Pap test is generally done for Cervical Cancer detection \cite{b15}. However, Ovarian Cancer has no proper prognosis method. Moreover, its status as the 7th most common cancer globally \cite{b12} for women makes it a dangerous disease for half the population of the world. In recent years, Computer Aided Detection (CAD) for diseases has become prevalent in the medical sector. From running simple blood tests to complex disease detection, machine learning has surely aided medical professionals by providing concise data as well as shortening diagnosis time for a disease or medical condition. Cancer is the most recent field where the application of machine learning has been seen \cite{b31}. A majority of cancer has early detection or testing methods with relevant involvement using machine language. However, ovarian cancer is more headache inducing when compared to the other forms of cancer disease as it has no early method of prognosis. Currently, the detection of ovarian cancer is done via a transvaginal ultrasound, a pelvic exam as well as CA-125 blood test \cite{b16}. However, a definitive result is only found via a lab-run biopsy for ovarian cancer \cite{b16}. Hence, researchers worldwide are attempting to find out new detection methods or improve the accuracy of already implemented machine learning models. Thus, introduction of an Artificial Intelligence that can accurately provide the resultant tumor type with minimal false report is an essential advancement.

In our study, we have selected an appropriate dataset for our base model training and testing and implemented tailored preprocessing techniques to optimize the data to our needs. We have utilized 15 variants of deep learning models based on LeNet, ResNet, VGG and GoogLeNet/Inception to select our base model and applied Explainable Artificial Intelligence (XAI) to explain the generated black box outcome.
\section{Related Works}
Zhou et al. published a research article \cite{b1} containing a review of the recent trends in the application of Artificial Intelligence in the field of diagnostic and prognostic prediction of ovarian cancer. They ended up with 39 studies that discussed the utilization of Artificial Intelligence in ovarian cancer and provided reasoning behind the larger number of high-throughput omic data that is the research trend on genomics and transcriptomes. They gave sound reasoning and reached the conclusion that the utilization of high-throughput data will increase not only in the field of cancer research but also in other medical sectors.

Hema et al. \cite{b2} presented a novel image classification model for ovarian cancer utilizing FaRe-ConvNN, which is a rapid region-based Convolutional neural network. In this model, they applied FaRe-ConvNN to perform the annotation procedure. The classification is done using a combination of SVC and Gaussian Naive Bayes classifiers after the region based training is completed. For testing the model, they utilized data from the Cancer Imaging Archive database. The researchers used epithelial cells, germ cells, and stromal cells samples separately. After comparing with existing models, the Gaussian Naive Bayes showed an accuracy score of 97\%, with sensitivity and specificity of 97.7\% and 98.69\% respectively. Based on the results, it can be concluded that the proposed model for ovarian cancer is an important contribution in the medical sector.

Wang et al. \cite{b3} developed a deep learning algorithm that can differentiate benign lesions from malignant lesions using magnetic resonance imaging in terms of ovarian cancer. The results showed that the model had higher accuracy and specificity against both the juniors (0.87 vs 0.64, 0.92 vs 0.64) and the seniors (0.87 vs 0.74, 0.92 vs 0.70). With the assistance of the model, the juniors showed a huge improvement in their accuracy (0.77 vs 0.64) and specificity (0.81 vs 0.64). In fact, the juniors showed higher specificity (0.81 vs 0.70) but similar accuracy (0.77 vs 0.74) while utilizing the proposed model when compared with the senior radiologists. In conclusion, the researchers said that the utilization of Artificial Intelligence can assist radiologists in assessing the nature of ovarian lesions while also improving their performance.

Another research article \cite{b4} proposed by Schwartz et al. featured an automated framework that detects ovarian cancer from transgenic mice using optical coherence tomography (OCT) recording. The basis of this proposal is the clear lack of noninvasive and viable source of early ovarian cancer prognosis. The researchers utilized three neural networks namely, a VGG-supported feed-forward network, a 3D CNN, and a convolutional Long Short-Term Memory (LSTM). Their experiments showed favorable results while LSTM showed the best AUC of 0.81 with a standard deviation of 0.037. They believe that the significance of this research lies in the fact that the usage of OCT can be a viable early prognosis for ovarian cancer.

A research paper by Hsu et al. \cite{b5} utilized ten convolutional neural network models for the detection and classification of ovarian cancer. They selected three  (ResNet-18, ResNet-50, and Xception) with the highest ratio of accuracy to time and utilized them for ensemble learning. Additionally, they used the interpretation of the ensemble classifiers as the result and visualized the decision making process using gradient-weighted class activation mapping (Grad-CAM) technology. For the database, they collected data from 587 patients from Taiwan following legal procedures. In their final discussion, they suggested that the confidence threshold be set at 80\%-100\% for the best possible outcome when using their model.

Another research paper proposed by Kasture et al. \cite{b7} is the first to identify, predict, and categorize ovarian cancer subtypes from histopathological images using VGG16. Initially, they trained the model with 500 images, 100 for each class, and obtained an accuracy of 50\%. They then multiplied the dataset of 500 images by doing several types of image augmentations to produce 24742 images. Then, they utilized this augmented image dataset to produce an accuracy of 84.64\%. Their core contributions are actually a wonderfully segmented image dataset that accurately classifies the various categories of ovarian cancer. Moreover, they displayed a series of accurate statistics that solidified their contribution of combining the prediction of ovarian cancer \& sub-type classification.
\section{Dataset}

Before the construction of a model for our work, we need to find an appropriate dataset that can be utilized to bring our future model to its full potential. As such, we selected the dataset “OvarianCancer\&SubtypesDatasetHistopathology” from Mendeley \cite{b8} to be the basis of our research. We picked this dataset since it not only has multiple different types of malignant tumor classes, it also has samples of non-cancerous classes as well.
\section{Methodology} 
\subsection{Data Pre-Processing}
\begin{figure}[h]
  \centering
  \includegraphics[width = .5\textwidth]{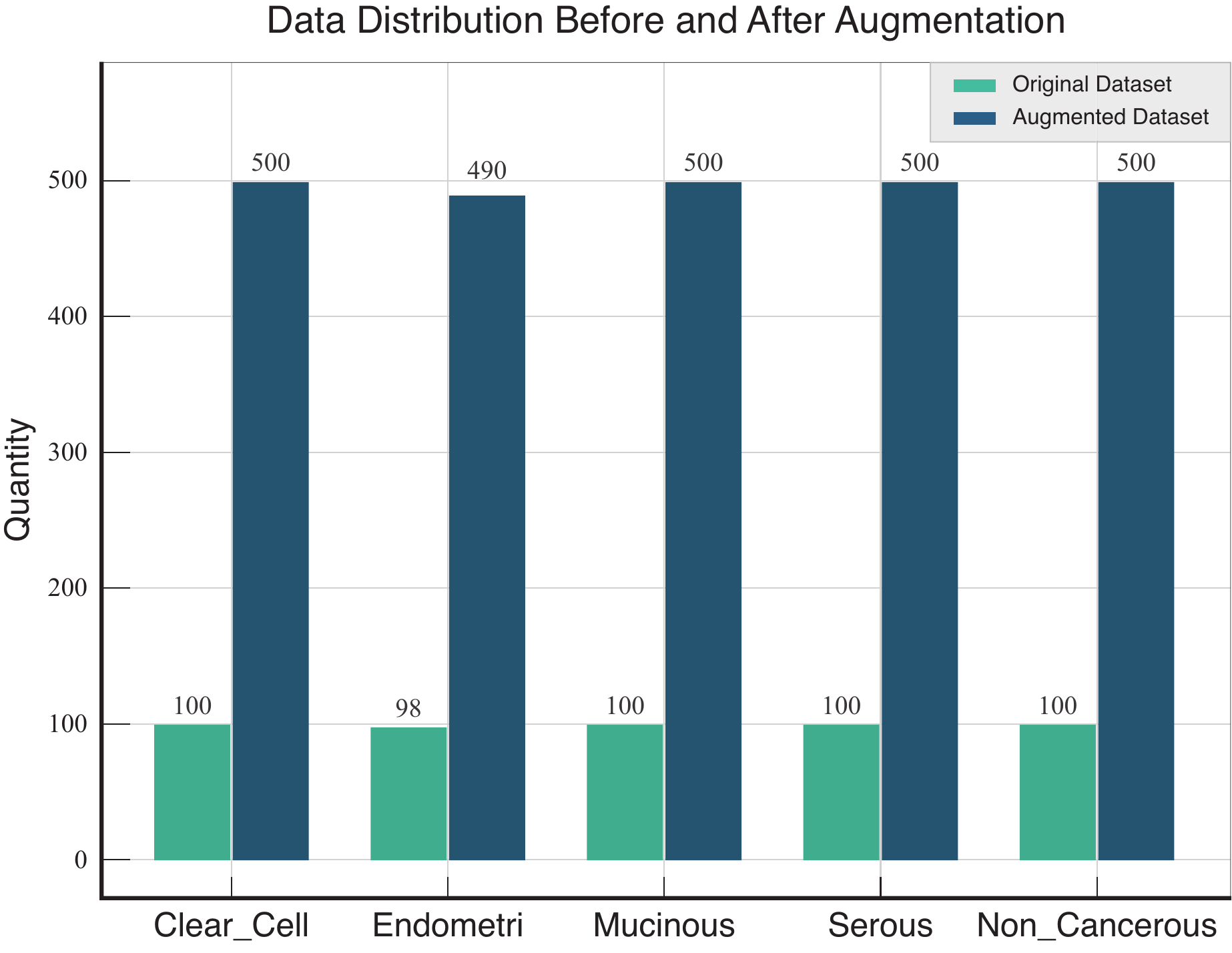}
  \caption{Data balancing bar chart featuring augmented images}
  \label{After augmentation}
\end{figure}

To prepare the dataset for the model, we needed to determine the features and perform appropriate pre-processing actions. First, to determine the features of the dataset, we ran a few tests and determined that the dataset we selected was balanced in nature as in the dataset, there were 98-100 images per class to a total of 498 images in 5 classes. Next, due to the limited size of the image dataset, we applied composite augmentation on the images using several types of transformations. These include: image rotation by up to 180 degrees, complete horizontal and/or vertical flipping, changes in brightness, contrast, saturation and hue to get 4 augmented images from each of the original images. To complete image augmentation, we used the Albumentations library and utilized various modules. The main reason for using the Albumentations library is because it has a variable probability in its transformations, indicating a greater percentage of randomness when augmenting any form of images. Furthermore, we ensured that the images follow the JPG image file format and that the color encoding of the images follow RGB. While performing the augmentation, we added the augmented and original images to a new sub-directory under our augmented dataset dictionary. Ultimately, our augmented dataset contained 5 subclasses with 2490 images in total which satisfied our requirements. The post-augmented data balance is seen in Figure-\ref{After augmentation}.

\subsection{Tensor Conversion and Value Normalization}
Next, we  converted the augmented images to tensor data using the image\_dataset \_from\_directory() method of the TensorFlow library. As we are performing a supervised learning approach for our model where the dataset directory we are passing in the method has subdirectories referring to the class names, we set the parameter “labels” to be “inferred”. Furthermore, we decided that one-hot encoding will become complicated if we decide to introduce more future subclasses. Hence, our “label\_mode” parameter was set to “int”. Table-\ref{table:2} explains the label and integer correlation. The color mode and batch size of the tensor data are set to RGB and 32 respectively. Our image size is variable in our initial testing, switching in between 32x32 and 224x224 per model requirements. Also, we initially split the dataset into a 80-20 ratio for training and testing. The 80-20 split was done randomly thanks to “seed”, “subset” and “shuffle” parameters. As a result, our tensor training dataset includes 1992 images while the tensor testing dataset contains 498 images. 

\begin{table}[h]
  \centering
  \captionsetup{justification=centering, labelsep=space}
  \caption{\textbf\\\vspace{-0.1em} Output Classification}
  \label{table:2}
  \renewcommand{\arraystretch}{1.25}
  \begin{tabular}{|>{\centering\arraybackslash}m{3cm}|>{\centering\arraybackslash}m{3cm}|}
  \hline
  \textbf{Sub-Class} & \textbf{Output Label} \\
  \hline
  \hline
  Clear Cell & 0 \\
  \hline
  Endometri & 1 \\
  \hline
  Mucinous & 2 \\
  \hline
  Non Cancerous & 3 \\
  \hline
  Serous & 4 \\
  \hline
  \end{tabular}
\end{table}

After completing the tensor conversion, we  decided to normalize the image dataset for a much smoother running experience with the various Convolutional Neural Network Models. Before doing that, the image portion of the tensor dataset has been converted from uint8 to float32 format using the tensorflow library. This is done so that the resultant scaling can be done much more easily. Only after doing so, we normalized the RGB values from a 0-255 range to a 0-1 range. We had tested this using some basic CNN structure and found that the later range provides a smoother convolution setup for the model. Lastly, we split the datasets, both training and testing into X and Y representing inputs and outputs. As our tensor conversion was done in batches, we used the concat() from tensorflow library method to create an input feature list and output label list for both training and testing datasets.

\subsection{CNN Models}
To select an appropriate model for accurate detection of different tumors from images, we aim to select an appropriate CNN model among the several variations of LeNet, ResNet, VGGNet and Inception. The base models and their variants are provided below. One thing we have utilized throughout most of our variant models is the usage of 'Softmax' activation function in the output layer \cite{b32}. The equation for 'Softmax' function is shown in Equation-(1):
\begin{align}
    \text{softmax}(z)_i = \frac{e^{z_i}}{\sum_{j=1}^{N} e^{z_j}}
\end{align}
\subsubsection{LeNet}
\begin{figure}[h!]
  \centering
  \includegraphics[width = \linewidth]{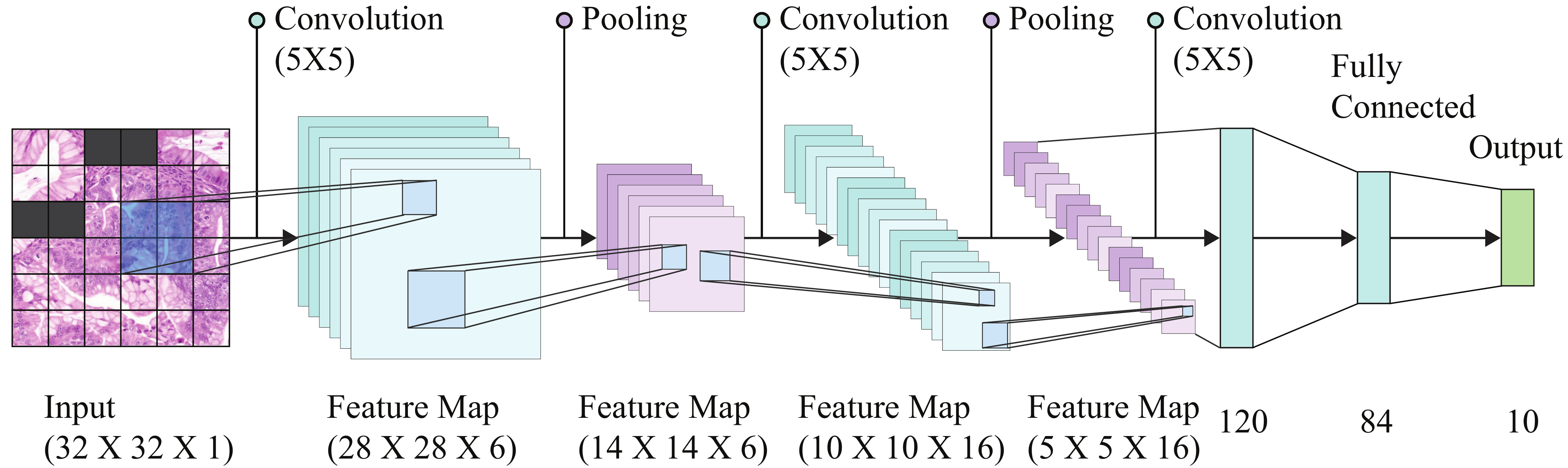}
  \caption{LeNet}
  \label{m_LeNet}
\end{figure}

The general model for LeNet \cite{b24} is quite simple in comparison to the other models we will be using \cite{b17}. There are three convolution layers, each utilizing 5x5 kernels with the filters being 6, 16, 120 for respectively. The output feature map from these convolution layers are 28x28x6, 10x10x16 and 120. After the first two convolution layers, a 2x2 max-pool is performed. After the final convolution layer, the nodes are flattened to ensure an easier time in constructing the output layer. For LeNet (Figure-\ref{m_LeNet}), we tested 8 variations and concluded that the following 3 were the better variants in terms of overall performance. All of the following models are evaluated over 100 epochs.
\begin{itemize}
    \item LeNet-A: The first LeNet is actually just the base model with a customized learning rate. Here, the learning rate was set to 0.001. 
    \item LeNet-B: This variant of LeNet takes the LeNet-A variant and adds in a dropout function to combat overfitting.
    \item LeNet-C: This variant of LeNet takes the LeNet-B variant and introduces step decay.
\end{itemize}

\subsubsection{ResNet}
The primary innovation of ResNet \cite{b9} is the introduction of residual connections which enables the network to learn residual mapping \cite{b21}. This effectively allows the connections to bypass one or more layers and propagate information directly to subsequent layers \cite{b21}, \cite{b22}. The output of the Residual Neural Network is determined by Equation-(2):
\begin{align}
    \hspace*{-2cm}
    y = F\left(x\right) + x
\end{align}
The skipping operations are done via two methods of signal propagation, namely, Forward Propagation and Backward Propagation. Equation-(3) shows the Forward Propagation for a single residual block:
\begin{align}
    \hspace*{-2cm}
    x_{n+1} = F\left(x_n\right) + x_n
\end{align}
Applying this recursively, we have Forward Propagation for multiple residual blocks shown in Equation-(4):
\begin{align}
    \hspace*{-2cm}
    x_L = x_l + \sum_{i=l}^{L-1} F\left(x_l\right)
\end{align}
where, $L$ is index of the last residual block and $l$ is that of any earlier block. This suggests that a signal is passed from a block $l$ to a deeper block $L$.

For Backward propagation, let us take the derivative of Forward Propagation with respect to $x_l$ and solve it as shown in Equation-(5):
\begin{align}
        \frac{\delta \epsilon}{\delta x_l} 
        &= \frac{\delta \epsilon}{\delta x_L} + \frac{\delta \epsilon}{\delta x_L} \times \frac{\delta}{\delta x_l} \sum_{i=l}^{L-1} F\left(x_l\right)
\end{align}
Here is the function where degradation has to be minimized. This suggests that a shallow signal $\frac{\delta \epsilon}{\delta x_l}$ has a term $\frac{\delta \epsilon}{\delta x_L}$ always added to it. Hence, the signal $\frac{\delta \epsilon}{\delta x_l}$ never disappears no matter how small the gradient of $F(x_l)$ becomes \cite{b23}.

\begin{figure}[ht]
  \centering
  \includegraphics[width = \linewidth]{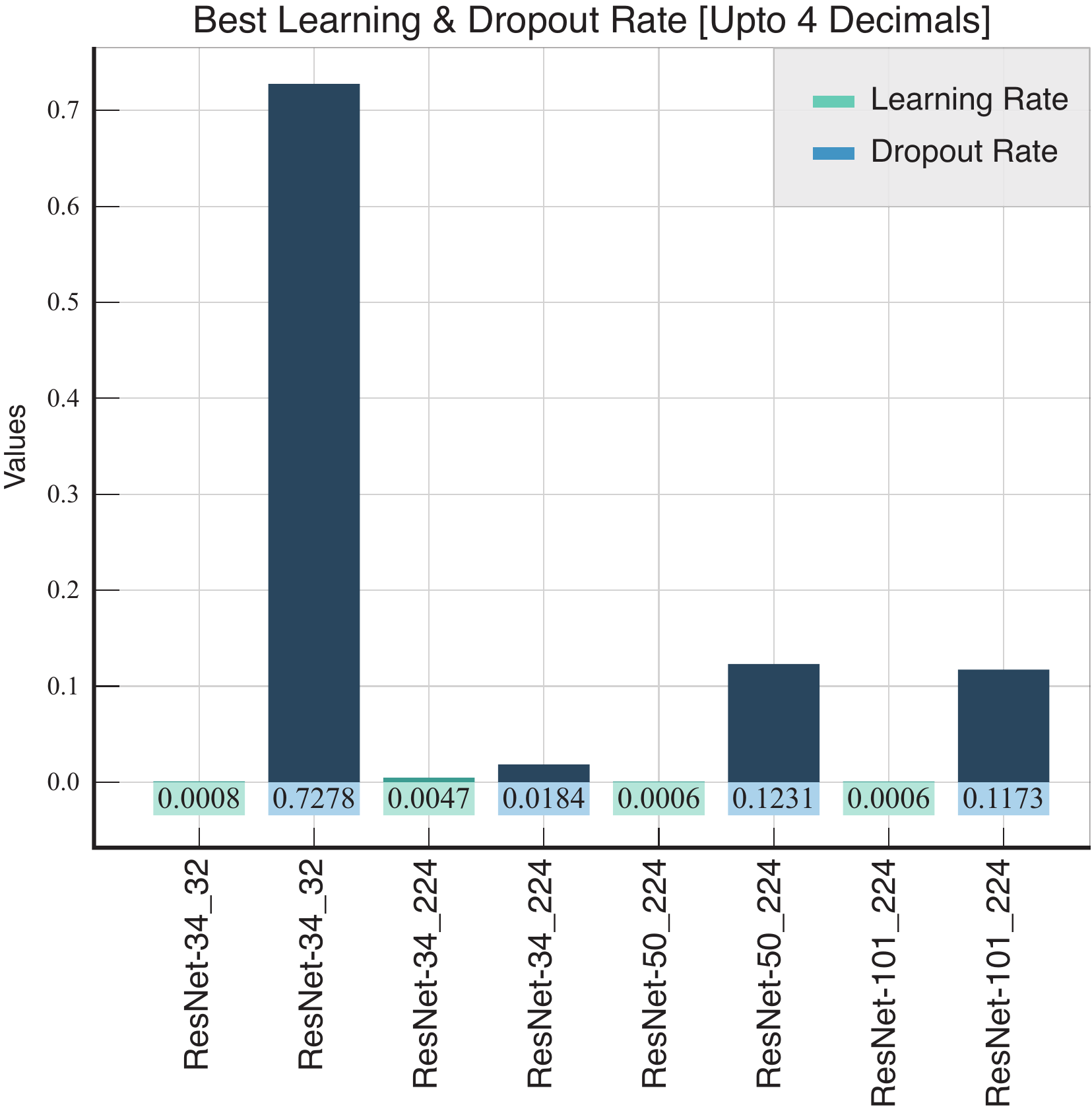}
  \caption{ResNet Best Learning Rate and Dropout rate tested over 30 iterations each}
  \label{ResNet best lr and dr}
\end{figure}

The basic building blocks of ResNet across its variations are quite similar \cite{b9}. We simply add residual connections every few 'blocks' or combinations of convolutional and maxpool layers. We will be primarily focusing on building ResNet-34 with two variable inputs of 32x32 and 224x224, ResNet-50 and ResNet-101 with 224x224 size image inputs. We optimized theses variants with some hyper-parameters that suits our needs such as utilizing learning rate and node dropouts for optimization control and avoiding overfitting. The various models each require a different ratio of learning rate and dropout rate. We cannot manually test each and every possible outcome as that would take an extremely long time. Thus, we took a randomized approach for every ResNet model. That is, we did the following for each variation of ResNet  models: We initially set the range for Learning Rate to be from 0.0001 to 0.1 and that of Dropout Rate from 0.0 to 0.9. We then took random sets of learning rate and dropout rate over 10 iterations and inserted the random hyperparameters in the model and ran over 3 epochs. At last, we selected the best learning rate and dropout rate based on the best testing accuracy. Our learning rate and dropout rate for all 4 of the variants are given in Figure-\ref{ResNet best lr and dr}.

\subsubsection{VGG}
The reason VGGNet \cite{b25} is called a deep CNN is because it has multiple layers with VGG-19 consisting of 19 convolutional layers and VGG-16 having 16 convolutional layers. As it is a very effective learning model, VGGNet is oftentimes used as a pre-trained model \cite{b26}.

\begin{figure}[h]
  \centering
  \includegraphics[width = 0.5\textwidth]{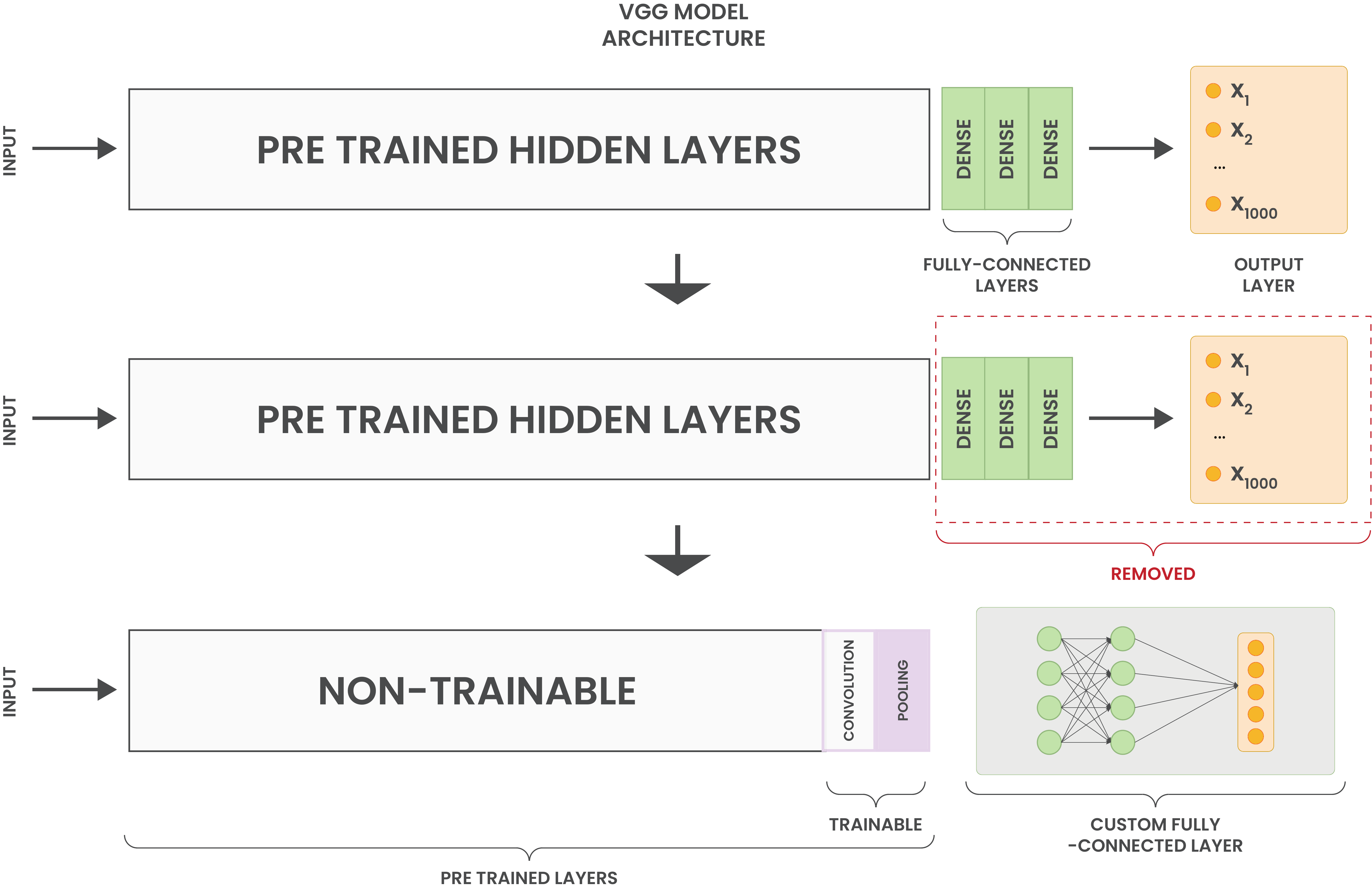}
  \caption{VGG Transfer learning process}
  \label{VGG_Architecture}
\end{figure}

We will be testing both VGG16 and VGG19 for the base model requirement. The variations in VGGNet will not matter much as we will be utilizing transfer learning. Compared to the other models, the heavy combination of consecutive convolution layers will result in extremely long training time that is also intensive in the aspect  of resource usage \cite{b18}. Thus, we will opt to use transfer learning in this case and use a pretrained base model with only the fully connected layer customized to our needs as seen in Figure-\ref{VGG_Architecture}.

\begin{itemize}
    \item VGG16-A: Here, we introduce a two dimensional global average pooling at the start of the fully connected layer. Next, we add three consecutive dense layers with ReLu activation functions with 1024, 1024 and 512 nodes respectively. Our output layer consists of 5 nodes and utilizes the softmax function.
    \item VGG16-B: The variant B is similar to variant A but it uses tanh activation function in the dense layers instead.
    \item VGG16-C: The variant C is similar to variant A but it adds learning rate of 0.03\% and dropout rate of 20\%.
    \item VGG19: Like the VGG16-A base model, we introduce a two dimensional global average pooling at the start of the fully connected layer. Next, we add three consecutive dense layers with ReLu activation functions with 1024, 1024 and 512 nodes respectively. Our output layer consists of 5 nodes and utilizes the softmax function.
\end{itemize}

\subsubsection{Inception}
\begin{figure}[h]
  \centering
  \includegraphics[width = .5\textwidth]{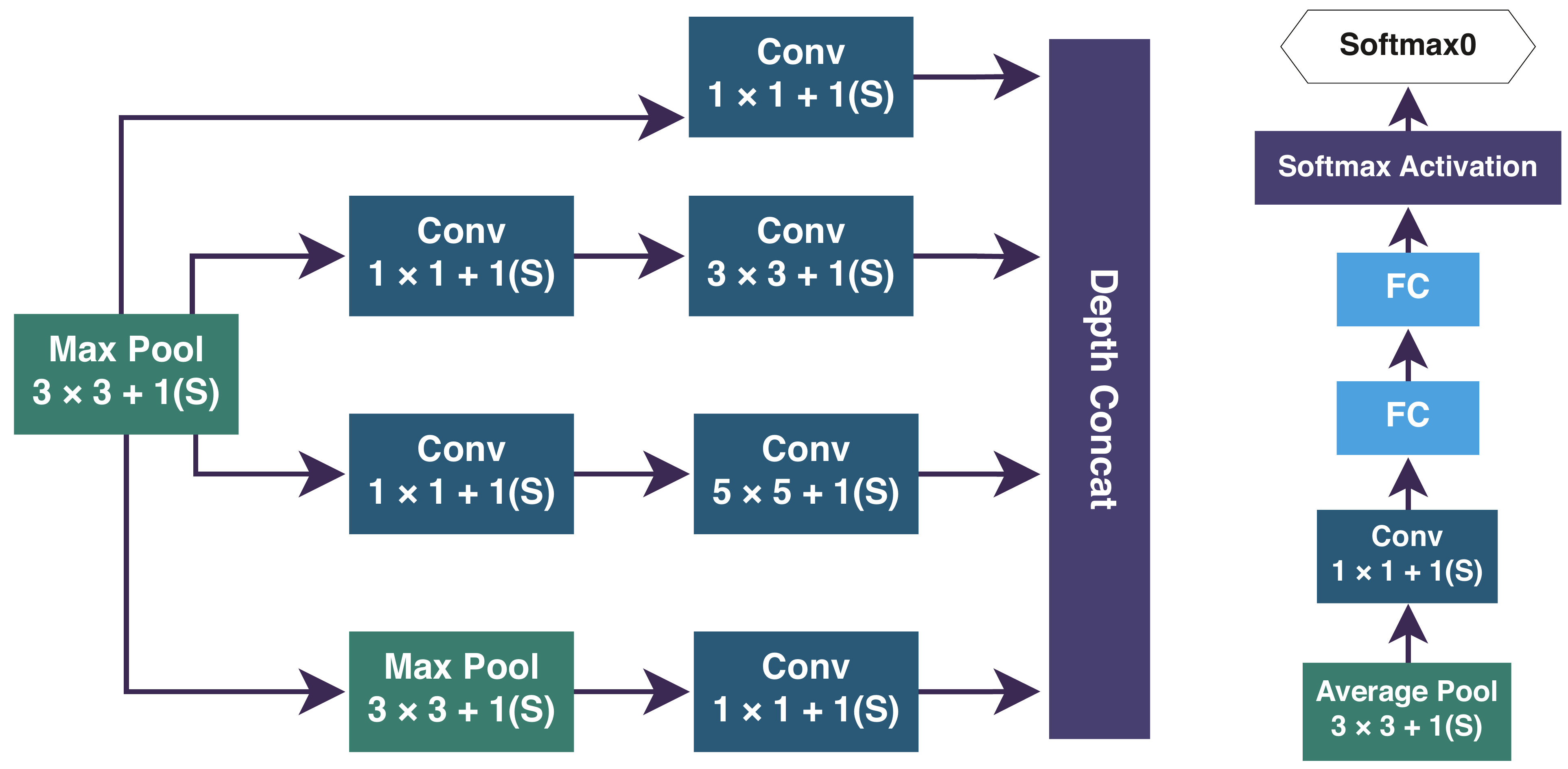}
  \caption{Inception Module (left) \& Auxiliary Classifier (Right)}
  \label{m_GoogleNet2}
\end{figure}

GoogLeNet, also known as InceptionNet \cite{b27}, has a total of 22 parameterized layers and 27 in total if including the non-parameterized layers such as the Max-Pooling layer \cite{b28}, \cite{b29}. GoogLeNet introduced Inception modules. These modules have kernels of sizes 1x1, 3x3 and 5x5 as seen in Figure-\ref{m_GoogleNet2}. The larger kernels cover greater area while the smaller ones cover the smaller but finer details in an image \cite{b29}. A total of 9 inception modules are used in GoogLeNet.

For GoogLeNet or Inception, we will be looking into Inception V1 and Inception V3 as these two versions are readily accessible. Like most of our other approaches, this model will be built from scratch. The  core mechanism of Inception is the usage of Inception modules. These modules utilize a series of 1x1, 2x2, 3x3, 5x5 convolution and maxpool layers to create a branching method such that the larger kernels cover the major details and the smaller ones  will cover the smaller details \cite{b14}. In both Inception V1 and V3, we will be using only two inception modules as that will be more than enough to tackle our chosen dataset. If we were to work with larger, complex datasets then we can opt to add in more inception modules and auxiliary classifiers according to our needs. The utilized variations are:
\begin{itemize}
    \item InceptionV1-A: We utilized a 2x2 average pooling before the final convolution and layer flattening. We also removed the auxiliary classifier. Furthermore, a generic single output layer is utilized instead of three. Lastly, the activation functions used were purely ReLu excluding the output layer.
    \item InceptionV1-B: InceptionV1-A's base is used here as well. The only difference here is that we utilized tanh activation function instead of ReLu.
    \item InceptionV3-A: We introduced batch normalization to the InceptionV1-A model. Here, 3 convolution layers have been utilized where the 1st layer is 3X3 instead of 7X7. Additionally, the filters of the inception modules have been modified where the 1st inception module has filters: 64, 128, 128, 32, 32, 32 and the 2nd inception module has filters: 128, 192, 96, 64, 64, 64.
    \item InceptionV3-B: InceptionV3-A's base is used here as well. The only difference here is that we utilized tanh activation function instead of ReLu.
\end{itemize}   
\subsection{XAI Models}
We have used 2 local post-hoc XAI, named LIME \cite{b33} and Integrated Gradients \cite{b34}, and 1 local variation of a global post-hoc XAI named SHAP \cite{b36} to explain the black box predictions of our selected AI model. Unlike the base models, we have not made any significant changes to the XAI models other than the visual representations.
\section{Result}
\subsection{Base model Results}
\begin{figure}[h!]
  \centering
  \includegraphics[width = 0.94\linewidth]{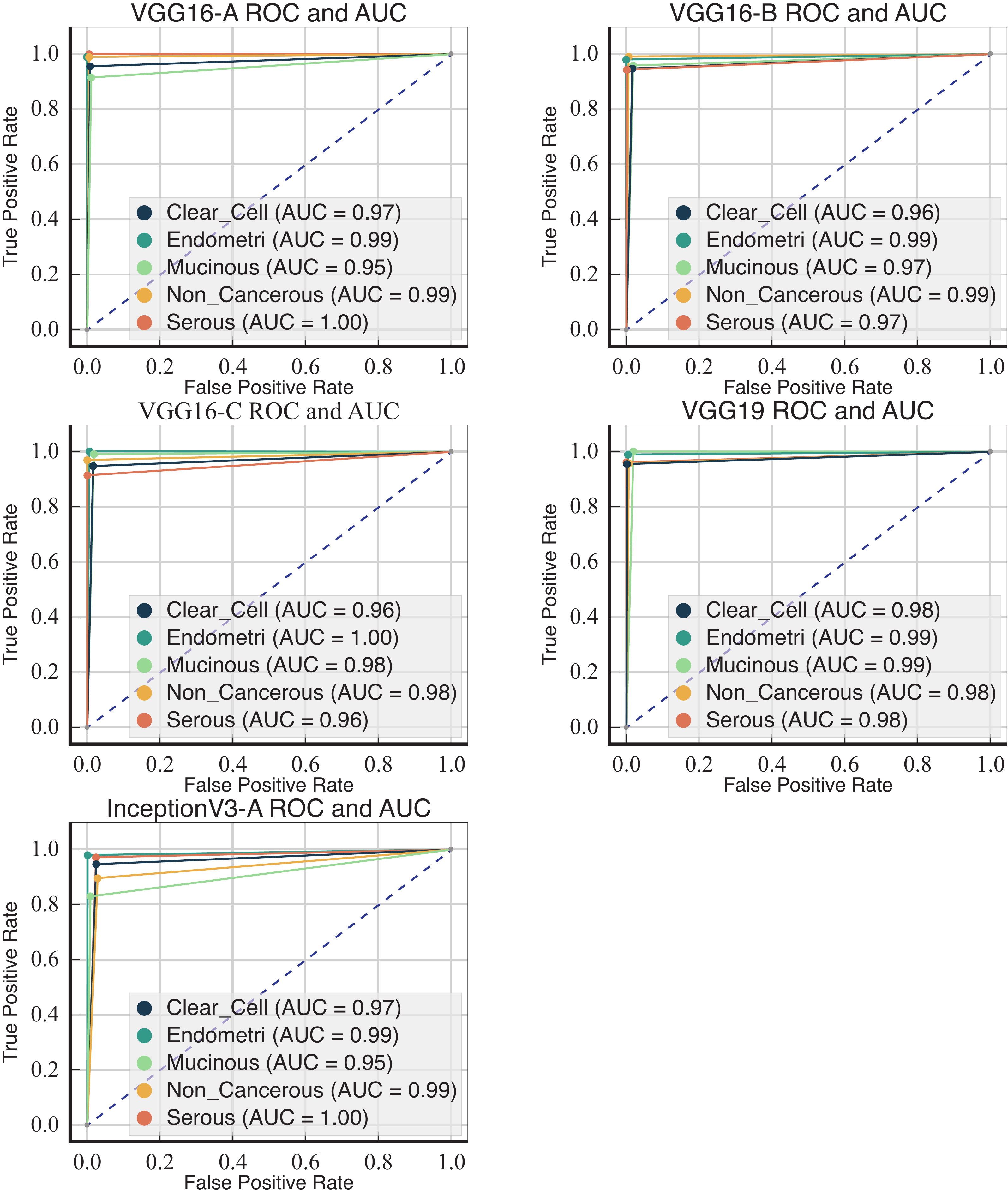}
  \caption{ROC and AUC}
  \label{roc-auc}
\end{figure}

From the results shown in Table-\ref{overall-result}, due to the scores of InceptionV3-A, VGG16 and VGG19 variants being the only ones above 90 across all four fields, we are showing their corresponding ROC Curves and AUC scores only in Figure-\ref{roc-auc}.
\vspace*{-5mm}
\begin{table}[h]
  \centering
  
  \captionsetup{justification=centering, labelsep=space}
  \caption{\textbf\\OVERALL RESULT}
  \label{overall-result}
  \renewcommand{\arraystretch}{1.25}
  \begin{tabular}{ |p{1.8cm}|p{1.075cm}|p{1.075cm}|p{1.075cm}|p{1.075cm}|  }
      \hline
      \textbf{Model Name} & \textbf{Accuracy} & \textbf{Precision} & \textbf{Recall} & \textbf{F1-Score} \\
      \hline
  \end{tabular}

  \vspace{-1mm}
  \begin{center}
      \textbf{LeNet}
  \end{center}
  \vspace{-1mm}

  \begin{tabular}{ |p{1.8cm}|p{1.075cm}|p{1.075cm}|p{1.075cm}|p{1.075cm}|  }
    \hline
    LeNet-A & 61.85\% & 62.20\% & 61.85\% & 61.96\% \\
    \hline
    LeNet-B & 55.02\% & 54.51\% & 55.02\% & 53.94\% \\
    \hline
    LeNet-C & 53.21\% & 55.28\% & 53.21\% & 49.53\% \\
    \hline
  \end{tabular}

  \vspace{-1mm}
\begin{center}
    \textbf{ResNet}
\end{center}
\vspace{-1mm}
\begin{tabular}{ |p{1.8cm}|p{1.075cm}|p{1.075cm}|p{1.075cm}|p{1.075cm}|  }

    \hline
    ResNet-34\_32 & 43.78\% & 36.67\% & 43.78\% & 38.30\% \\
    \hline
    ResNet-34\_224 & 57.03\% & 59.39\% & 57.03\% & 57.70\% \\
    \hline
    ResNet-50 & 34.14\% & 47.75\% & 34.14\% & 33.47\% \\
    \hline
    ResNet-101 & 43.17\% & 47.17\% & 43.17\% & 40.64\% \\
    \hline
\end{tabular}

\vspace{-1mm}
\begin{center}
    \textbf{VGG}
\end{center}
\vspace{-1mm}
\begin{tabular}{ |p{1.8cm}|p{1.075cm}|p{1.075cm}|p{1.075cm}|p{1.075cm}|  }

    \hline
    VGG16-A & 96.99\% & 96.98\% & 96.99\% & 96.97\% \\
    \hline
    VGG16-B & 96.18\% & 96.27\% & 96.18\% & 96.20\% \\
    \hline
    VGG16-C & 96.18\% & 96.32\% & 96.18\% & 96.18\% \\
    \hline
    VGG19 & 97.19\% & 97.31\% & 97.19\% & 97.20\% \\
    \hline
\end{tabular}

\vspace{-1mm}
\begin{center}
    \textbf{Inception}
\end{center}
\vspace{-1mm}
\begin{tabular}{ |p{1.8cm}|p{1.075cm}|p{1.075cm}|p{1.075cm}|p{1.075cm}|  }

    \hline
    InceptionV1-A & 78.92\% & 81.58\% & 78.92\% & 79.33\% \\
    \hline
    InceptionV1-B & 85.74\% & 86.26\% & 85.74\% & 85.42\% \\
    \hline
    InceptionV3-A & 94.58\% & 94.75\% & 94.58\% & 94.62\% \\
    \hline
    InceptionV3-B & 82.13\% & 85.11\% & 82.13\% & 82.70\% \\
    \hline
\end{tabular}

\end{table}

\subsection{XAI Result}
Due to space constraints, we are showing only a single result from LIME (Figure-\ref{LIME_clear_cell_0}) and Integrated Gradients (Figure-\ref{IG_clear_cell_0}) while we are showing the entire generated subplot for SHAP (Figure-\ref{SHAP}).
\begin{enumerate}
    \item LIME:
\begin{figure}[h]
  \centering
  \includegraphics[width = 0.9\linewidth]{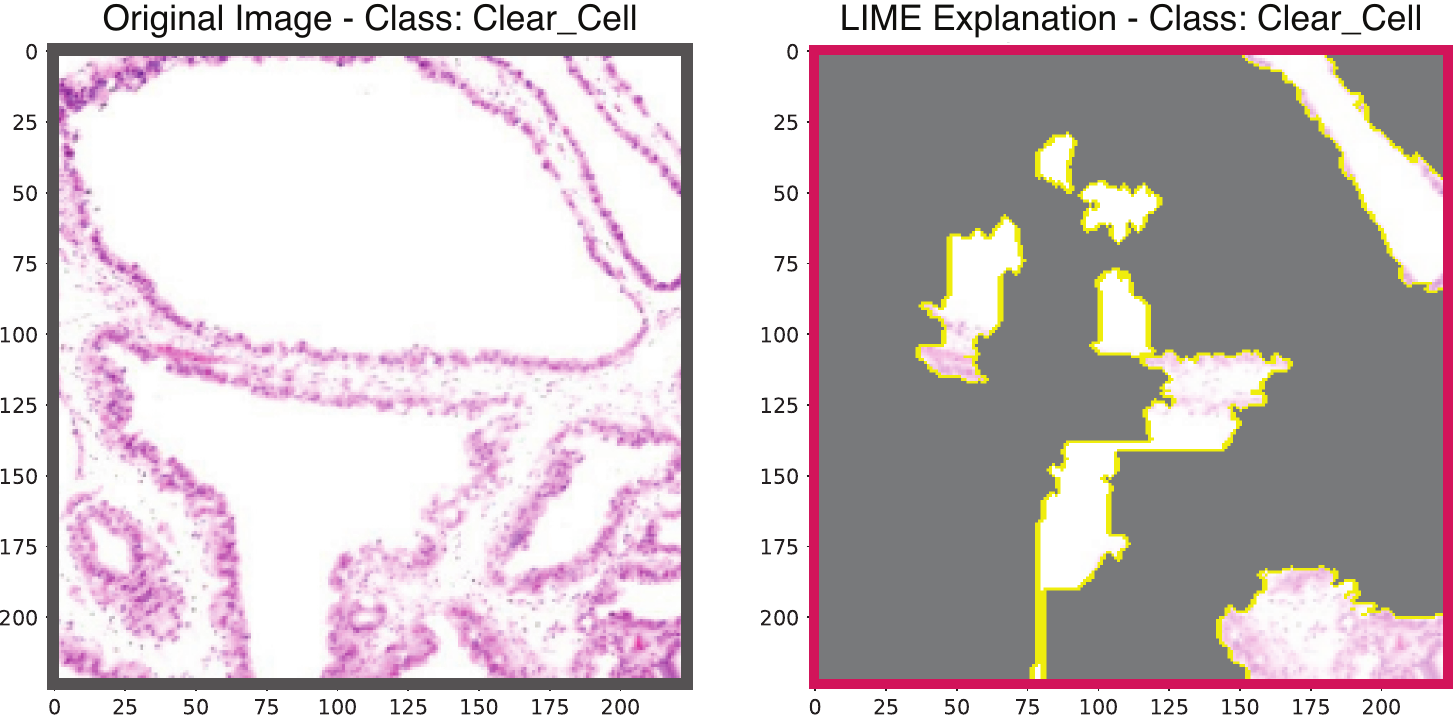}
  \caption{LIME (Class: Clear Cell)}
  \label{LIME_clear_cell_0}
\end{figure}

    \item Integrated Gradients:
\begin{figure}[h]
  \centering
  \includegraphics[width = \linewidth]{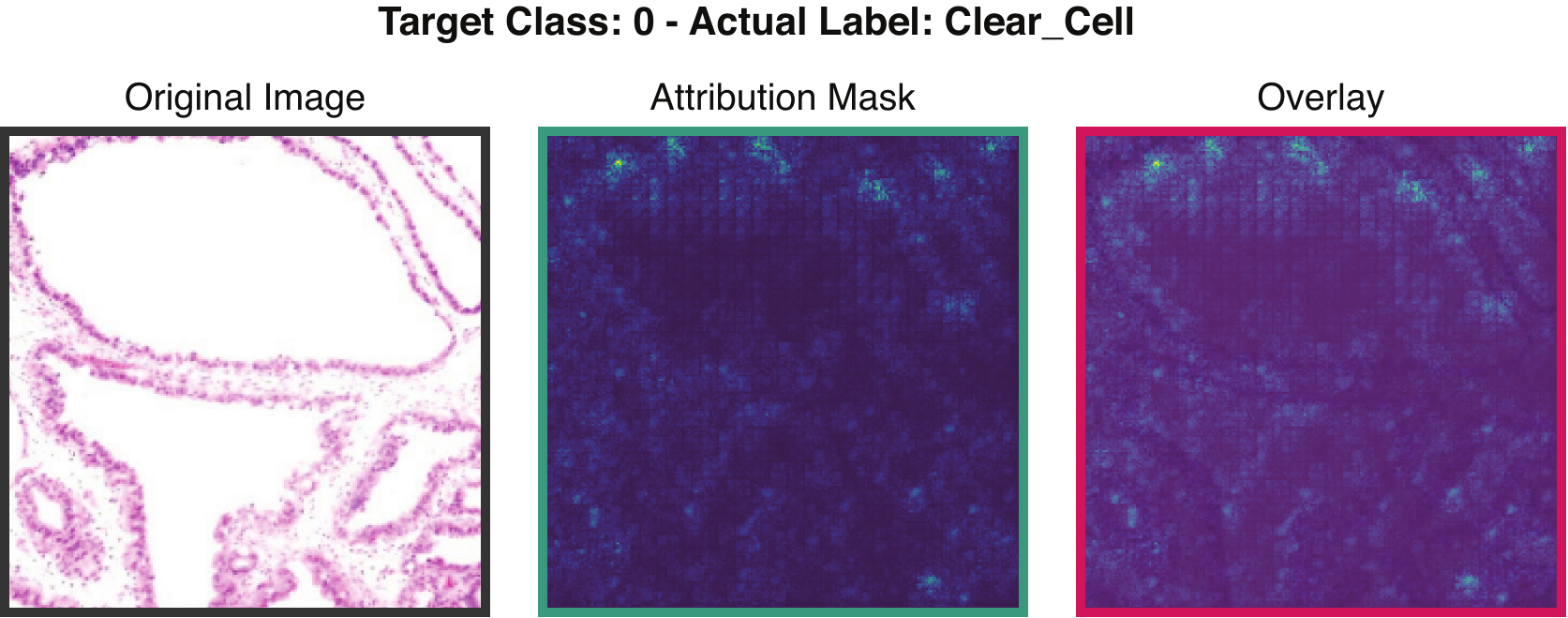}
  \caption{Integrated Gradients (Class: Clear Cell)}
  \label{IG_clear_cell_0}
\end{figure}

    \item SHAP:
\begin{figure}[h!]
  \centering
  \includegraphics[width = .97\linewidth]{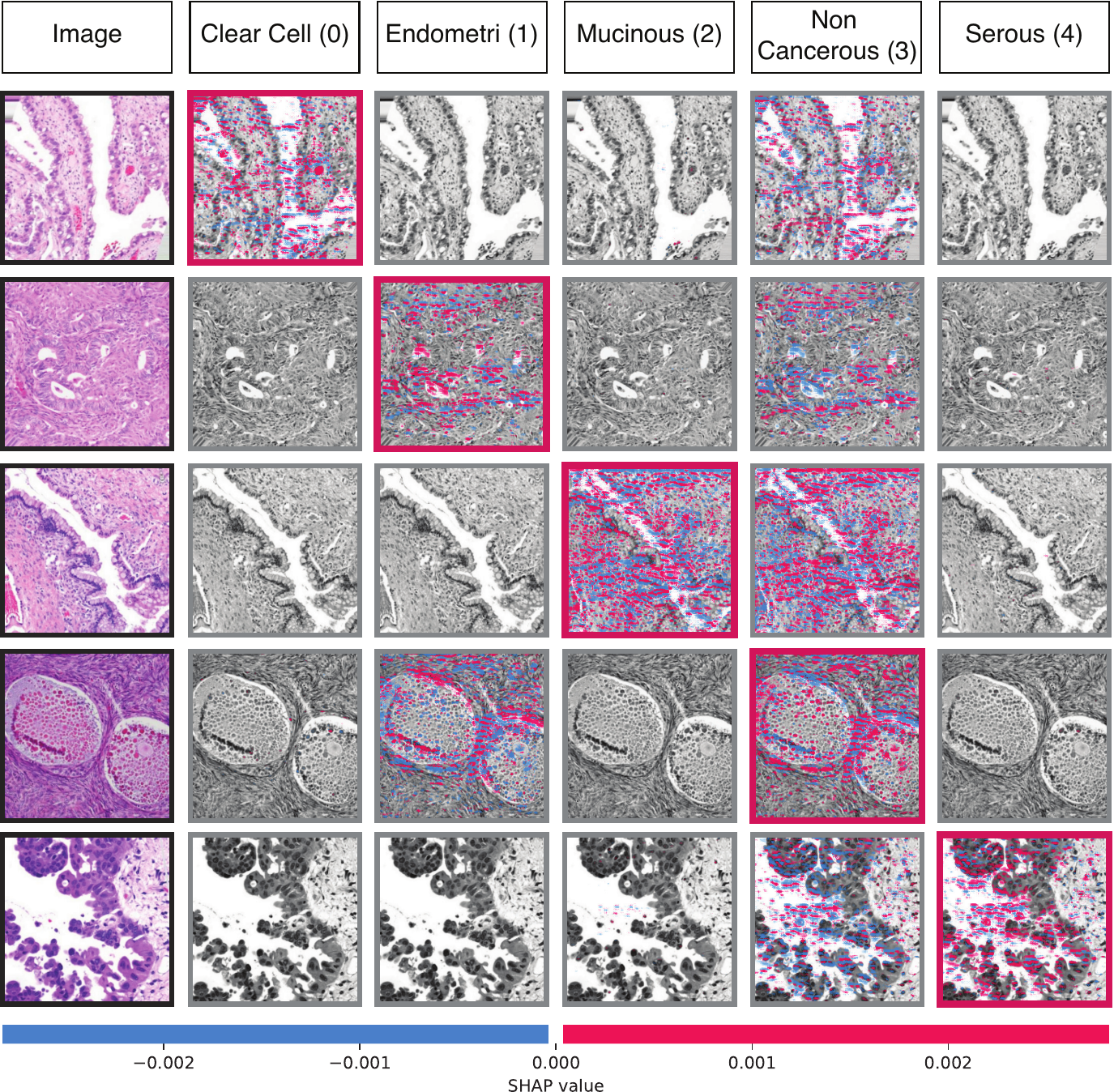}
  \caption{SHAP (Local)}
  \label{SHAP}
\end{figure}    
\end{enumerate}
\section{Analysis}
\subsection{Base Model: Inception V3-A}
There is a reason why InceptionV3-A has been selected and not the other models that were better performing. Let us first come to terms with the problems that will occur if VGG models were to be used. After the selection of a base model, we needed to work with explainable artificial intelligence or XAI to comprehend the black box answer that is produced via our selected model. The core of transfer learning makes it so that the utilization of XAI on models made via transfer learning is tremendously difficult when compared to that of a model that is built from scratch. Hence, the VGG models were rejected despite their high scores. Now, the model with the next highest score is InceptionV3-A. Thus, ultimately, our choice of model is the custom Inception V3 with ReLu activation function.

\subsection{Model Comparison}

\begin{table}[h]
  \centering

    \captionsetup{justification=centering, labelsep=space}
    \caption{\textbf\\Comparison of Average Model Accuracy between two of our models and one predecessor model}
    \label{table:4}
  \renewcommand{\arraystretch}{1.25}
  \begin{tabular}{|c|c|c|c|} \hline
    \textbf{Model}& \textbf{VGG16-O}\cite{b7}& \textbf{VGG16-A}&\textbf{InceptionV3-A}\\ \hline  \hline
       Original Dataset&  50\%& 77.78\%&20.20\%\\ \hline  
       Augmented Dataset& \makecell{84.64\%\\(20 epoch,\\24742 images)}& \makecell{96.99\%\\(80 epoch,\\2490 images)} &\makecell{94.58\%\\(80 epoch,\\2490 images)} \\ \hline 
  \end{tabular}
\end{table}

\noindent{Another thing that has been tested was our model score with the model of another paper by Kasture et al. that utilized the same dataset\cite{b7} (Henceforth, referred to as VGG16-O). According to Table-\ref{table:4},  VGG16-O achieved a score of 50\% with the non-augmented dataset. We also ran a minor test with our models VGG16-A and InceptionV3-A by running them for 20 epoch under our original  conditions. The average accuracy achieved was 27.78\% higher than that of VGG16-O. This can be attributed to the fact that Tensor Conversion had been performed after image augmentation. By converting the images to Tensor Data and further normalizing the values in a range of 0 to 1,  enabled for further computational efficiency and easier training for the base models. As for InceptionV3-A, the reason that it is performing much worse than VGG16-O and VGG16-A for the original dataset is because it is not pre-trained like the other models.}

\subsection{XAI Comparative Analysis}
From Figure-\ref{SHAP}, it has been observed that the first image belonging to “clear cell” class and find that it has more positive correlation associated with the “clear cell” category in comparison to other classes. Similarly, upon examining the remaining images, a consistent dominance of positive features aligned with their respective actual target classes can be noticed, providing a solid and precise rationale for its specific classification.

\begin{figure}[h]
  \centering
  \includegraphics[width = \linewidth]{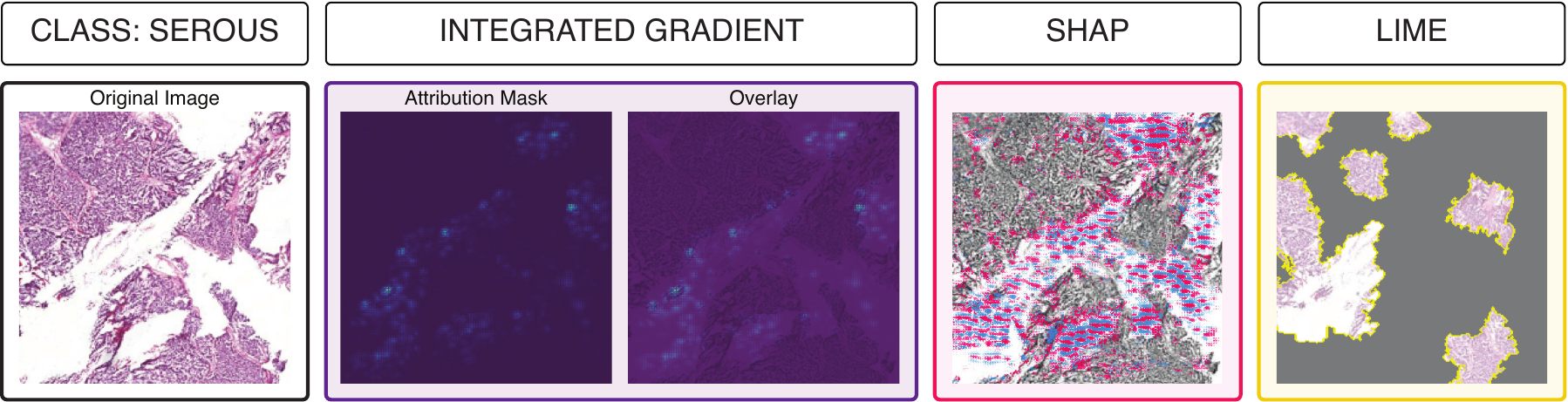}
  \caption{Comparative Analysis of Generated XAI outputs}
  \label{comp_analysis}
\end{figure}

In Figure-\ref{comp_analysis}, local visualized interpretation from LIME, SHAP and Integrated Gradients indicate that all three interpretations have similar highlighted features that contribute to the prediction of Serous Class. However, the reason why some highlighted features from SHAP and Integrated Gradients do not exist in LIME is that, LIME interpretation has been capped to showing only 10 important features to reduce complexity in analysis.

\section{Conclusion}
Cancer is a highly invasive disease that forms due to the abnormal growth of cells in any part of the body. Ovarian cancers are considered to be more deadly than other common cancers among women because of its late-stage prognosis. A late stage prognosis often means a high risk of the cancer cells spreading to other organs and thus increasing the chance of mortality. In the United States of America, ovarian cancer is deemed as the deadliest gynecologic cancer. Due to its high lethality, researchers all over the world are attempting to find either a faster and accurate detection method or a non-invasive detection method. In this paper, an automated detection system has been created that utilizes Convolutional Neural Networks (CNN) to detect ovarian cancer fast and accurately. For building such a system, different CNN models such as LeNet-5/LeNet, Residual Neural Network (ResNet), VGGNet and GoogLeNet/Inception have been utilized. After testing various iterations of the CNN models, Inception V3 has been used as the base AI for this endeavor. Explainable Artificial Intelligence (XAI) models such as Local Interpretable Model-agnostic Explanations (LIME), SHapley Additive exPlanations (SHAP) and Integrated Gradients has also been implemented for this system so that the outcome of the system can be interpreted and judged accordingly. Ultimately, a great initial success has been achieved by building a sandbox InceptionV3 model with the selected model that achieved an average score of 94.5\% to 94.75\% in the performance metrics such as Accuracy, Precision, Recall, F1-Score, ROC Curve and AUC. Moreover, the model also had one of the better ROC Curves and AUC scores when compared to the other 14 variations of the different CNN models that were experimented with. Next, a Comparative Analysis has been performed on the generated output of three different XAI models namingly LIME (Local Interpretable Model Agnostic Explanations), SHAP (SHapley Additive exPlanations) and Integrated Gradients with the results indicating that the generated outputs had some highlighted features that were common across the 3 models. This signifies that the black-box interpretation occurred successfully. Thus, it can be noted that an initial step was taken towards completing a system that can provide either an accurate, faster detection model or an early prognosis model. In the future, we aim to streamline the system and pivot towards early prognosis and faster detection by using non-invasive data as our image dataset.

\vspace{12pt}

\end{document}